\documentclass[conference]{IEEEtran}
\IEEEoverridecommandlockouts
\usepackage{cite}
\usepackage{amsmath,amssymb,amsfonts}
\usepackage{algorithm}
\usepackage{graphicx}
\usepackage{algpseudocode}
\usepackage{textcomp}
\usepackage{xcolor}
\usepackage{tcolorbox}
\tcbuselibrary{skins}
\usepackage{lipsum}
\usepackage{booktabs}
\usepackage{bbding}
\usepackage{tikz}
\usepackage{wrapfig}
\usepackage{url}
\usepackage{footmisc}
\newcommand{\mb}[1]{{\mathbb{#1}}}

\newcommand{\algorithmicline}{\rule{\linewidth}{0.5pt}}

\newcommand{\equalcontribfootnote}[1]{%
  \begingroup
  \renewcommand{\thefootnote}{}% Remove footnote numbering
  \footnotetext{#1}%
  \endgroup
}

\newcommand{\equalcontrib}{\textsuperscript{*}}

\def\BibTeX{{\rm B\kern-.05em{\sc i\kern-.025em b}\kern-.08em
    T\kern-.1667em\lower.7ex\hbox{E}\kern-.125emX}}
\begin{document}
\title{SEFD: Semantic-Enhanced Framework for Detecting LLM-Generated Text\\
}

\author{\IEEEauthorblockN{Weiqing He\equalcontrib{}}
\IEEEauthorblockA{\textit{School of Art and Science} \\
\textit{University of Pennsylvania}\\
Philadelphia, USA \\
weiqingh@sas.upenn.edu}
\and
\IEEEauthorblockN{Bojian Hou\equalcontrib{}}
\IEEEauthorblockA{\textit{Department of Biostatistics}, \\ \textit{Epidemiology and Informatics} \\
\textit{University of Pennsylvania}\\
Philadelphia, USA \\
bojian.hou@pennmedicine.upenn.edu}
\and
\IEEEauthorblockN{Tianqi Shang\equalcontrib{}}
\IEEEauthorblockA{\textit{Department of Biostatistics}, \\ \textit{Epidemiology and Informatics} \\
\textit{University of Pennsylvania}\\
Philadelphia, USA \\
tianqi.shang@pennmedicine.edu}
\and
\IEEEauthorblockN{Davoud Ataee Tarzanagh}
\IEEEauthorblockA{\textit{Department of Biostatistics}, \\ \textit{Epidemiology and Informatics} \\
\textit{University of Pennsylvania}\\
Philadelphia, USA \\
davoud.ataeetarzanagh@pennmedicine.upenn.edu}
\and
\IEEEauthorblockN{Qi Long}
\IEEEauthorblockA{\textit{Department of Biostatistics}, \\ \textit{Epidemiology and Informatics} \\
\textit{University of Pennsylvania}\\
Philadelphia, USA \\
qlong@pennmedicine.upenn.edu}
\and
\IEEEauthorblockN{Li Shen$^\dagger$\thanks{$\dagger$ Corresponding author}}
\IEEEauthorblockA{\textit{Department of Biostatistics}, \\ \textit{Epidemiology and Informatics} \\
\textit{University of Pennsylvania}\\
Philadelphia, USA \\
li.shen@pennmedicine.upenn.edu}
}

\maketitle

% \footnotetext[0]{\textsuperscript{*}These authors contributed equally to this work.}
\equalcontribfootnote{* These authors contributed equally to this work.}

\begin{abstract}
% The increasing use of large language models (LLMs) highlights the need for enhanced tools to detect LLM-generated text. Paraphrasing presents a significant challenge, often bypassing detection and reducing the effectiveness of existing tools. To address this, we introduce a novel framework for detecting LLM-generated text that incorporates a retrieval pool to fully leverage text semantics. This approach markedly enhances the performance of existing detection methods, including the soft watermarking strategy.

The widespread adoption of large language models (LLMs) has created an urgent need for robust tools to detect LLM-generated text, especially in light of \textit{paraphrasing} techniques that often evade existing detection methods. To address this challenge, we present a novel semantic-enhanced framework for detecting LLM-generated text (SEFD) that leverages a retrieval-based mechanism to fully utilize text semantics. Our framework improves upon existing detection methods by systematically integrating retrieval-based techniques with traditional detectors, employing a carefully curated retrieval mechanism that strikes a balance between comprehensive coverage and computational efficiency. We showcase the effectiveness of our approach in sequential text scenarios common in real-world applications, such as online forums and Q\&A platforms. Through comprehensive experiments across various LLM-generated texts and detection methods, we demonstrate that our framework substantially enhances detection accuracy in paraphrasing scenarios while maintaining robustness for standard LLM-generated content. This work contributes significantly to ongoing efforts to safeguard information integrity in an era where AI-generated content is increasingly prevalent. Code is available at \url{https://github.com/hwq0726/SEFD}.
\end{abstract}

\begin{IEEEkeywords}
LLMs, LLM-Generated Text Detection, Paraphrasing, Semantic Analysis, Information Retrieval
\end{IEEEkeywords}

\section{Introduction}
% The emergence and evolution of Large Language Models (LLMs) such as GPT-3 \cite{brown2020language}, PaLM \cite{chowdhery2023palm}, and ChatGPT \cite{chatgpt} mark a transformative phase in the field of artificial intelligence. These models, with their unprecedented scale and complexity, have demonstrated remarkable versatility, enabling a multitude of applications in various domains. From enhancing creative processes to improving predictive analytics, these LLMs have not only pushed the boundaries of machine learning but also catalyzed a paradigm shift in human-computer interaction. Their ability to understand and generate human-like text has opened new frontiers in natural language processing, making them indispensable tools in modern technology.

The proliferation of LLM-generated content, while beneficial in many ways, also raises significant concerns, especially in terms of the authenticity and accuracy of information. Instances of misleading or incorrect LLM-generated text, as evidenced in the study by \cite{lin2021truthfulqa}, highlight the imperative need for effective detection mechanisms. Additionally, in human health-related research, the authenticity of data is paramount~\cite{he2024interpretability, shang2024leveraging}, as reliance on fabricated or inaccurately generated data, such as AI-generated medical notes, could fundamentally compromise study outcomes. Ensuring data integrity is essential to maintain trustworthiness~\cite{hou2020learning,hou2017learning}, especially when findings directly impact patient care or public health policies~\cite{hou2024interpretable}.

Innovations in this field include watermarking \cite{kirchenbauer2023watermark,abdelnabi2021adversarial,grinbaum2022ethical}, training based classifiers \cite{bhattacharjee2024eagle, hovy2016enemy,zellers2019defending,bakhtin2019real,ippolito2019automatic,openaiclassifier}, statistical test based detectors \cite{mitchell2023detectgpt, gehrmann2019gltr, solaiman2019release, ippolito2019automatic} and text heterogeneity based detectors \cite{tulchinskii2024intrinsic, yang2023dna, mao2024raidar}. In an era where LLM-generated content is increasingly dominant, these tools represent a concerted effort to ensure the integrity and reliability of information.

Identifying LLM-generated text poses a significant challenge due to the rapid evolution of LLMs and widespread evasion tactics \cite{sadasivan2023can}. Concurrently, evasion techniques such as \textit{paraphrasing}—where another LLM is used to rephrase the original LLM-generated text—undermine the effectiveness of detection tools like GPTZero \cite{zerogpt}, DetectGPT \cite{mitchell2023detectgpt}, and OpenAI's text classifier \cite{openaiclassifier}. These methods are non-watermarking techniques that rely on LLM-specific features. Paraphrasing can obscure these features by altering text attributes, which hampers detection. Furthermore, watermarking strategies that insert unique patterns into LLM outputs are vulnerable to evasion through paraphrasing \cite{krishna2023paraphrasing} which can eliminate the markers by restructuring sentences and replacing synonyms.

% \hbj{The logic of this paragraph could be in this way: To tackle the paraphrasing issue, we introduce a retrieval-based method that utilizes the similarity of semantic meaning between the target text and the AI texts from a pre-defined database to detect whether the target text is AI generated or not. 
% By systematically combining such technique with the traditional detection method, there is chance to not only gain a good detection performance for regular AI text but also handle the paraphrasing issue well. However, the effectiveness of the retrieval-based technique relies on the comprehensiveness of its database...To tackle these challenges, we propose a framework that is able to not only systematically combine the retrieval-based method with the traditional method but also maintain a database with reasonable size...}
To address the paraphrasing issue, we introduce a semantic similarity-based retrieval technique that computes the semantic similarity between the target text and LLM-generated texts from a pre-defined database to determine whether the target text is LLM-generated. Since paraphrasing largely preserves the original semantic meaning, this technique offers a viable approach to defend against paraphrasing attacks. By systematically integrating this technique with traditional detection methods, we aim to achieve robust detection performance for both regular LLM-generated text and paraphrased content. However, the effectiveness of the retrieval technique heavily depends on the comprehensiveness of its database, which presents significant practical challenges. On one hand, storing all LLM responses demands significant storage space and computational resources for retrieval from this extensive dataset. On the other hand, to save storage, a limited database may not be sufficient and can be outdated as the environment or distribution changes.
Given the challenges posed by paraphrasing tactics and the limitations of the retrieval technique, we propose a framework that is able to not only systematically combine it with the traditional method but also update in real-time as detection progresses.
\begin{figure}[t]
  \centering
  \includegraphics[width=9cm]{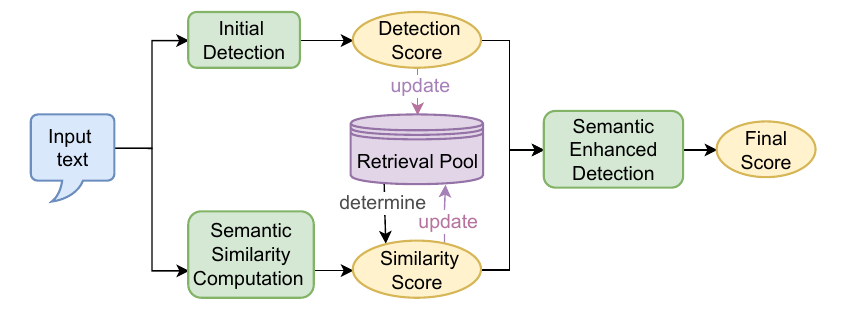}
  \caption{A brief version of SEFD structure. SEFD comprises a retrieval database/pool (colored by purple) and three detection steps (colored by green).}
  \label{fig:simpleflow}
\end{figure}

\begin{figure*}[h]
  \centering
  \includegraphics[width=16cm]{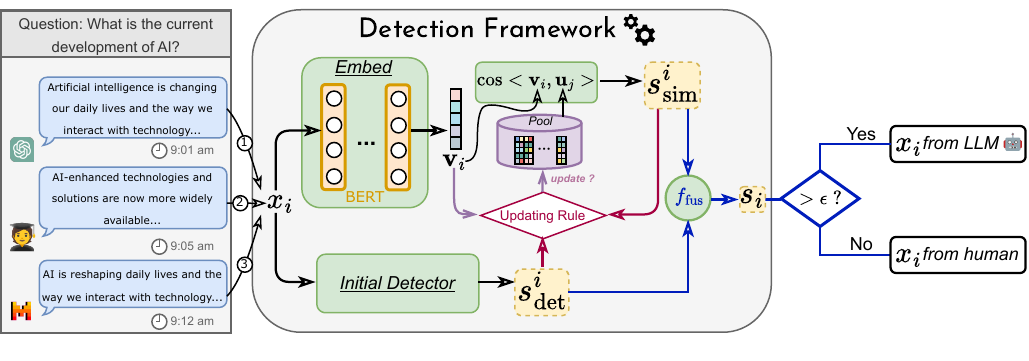}
  \vspace{-0.4cm}
  \caption{The detailed structure of our framework. The input sequence on the left consists of three texts: the first is generated by an LLM, the second is human-written, and the third is a paraphrased version of the first text by another LLM. These texts are processed in order. For text $x_i$, we conduct initial detection and semantic similarity computation simultaneously to get detection score and similarity score. These scores are then combined using the fusion function to produce the semantic-enhanced detection score, which classifies $x_i$. Finally, based on an updating rule, we decide whether to include $x_i$ in the retrieval pool and proceed to detect the next text, $x_{i+1}$.}
  \label{fig:flow}
\end{figure*}

% We first define the \textbf{Input Text Assumptions} with three types of input: LLM-generated text (paraphrased version excluded), paraphrased LLM-generated text, and human-written text. 
%\hbj{We don't need to introduce this detailed assumption in the introduction}
%We assume that inputs are strictly sequential; no simultaneous inputs are allowed, and for LLM-generated text, its paraphrased version always follows the original. For example, let “$a$”, “$b$”, and “$c$” represent LLM-generated texts, and “$\hat{a}$”, “$\hat{b}$”, and “$\hat{c}$” represent their paraphrased versions. A valid input sequence would be: “$a;b;\hat{a};\hat{b};c;\hat{c}$.”
% To illustrate the assumptions, here is an example of the input text sequence.
% \begin{quote}
% \vspace{-0.1cm}
%     \textbf{Example}. we use ``$a$'', ``$b$'' and ``$c$'' to represent three different LLM-generated texts and use ``$\hat{a}$'', ``$\hat{b}$'' and ``$\hat{c}$'' to denote the paraphrased version respectively then, ``\textit{$a\;b\;\hat{a}\;\hat{b}\;c\;\hat{c}$}''  can be input in left-to-right order.
% \vspace{-0.3cm}
% \end{quote}
%
% In practice, as we mentioned above, there are many scenarios that closely align with these assumptions, like detecting the source of comments under a post or detecting the source of answers under a question.
%These assumptions are applicable in scenarios such as detecting the source of comments or answers in online posts and questions.

We also observe that in many detection scenarios, texts follow a sequential order, such as detecting the source of comments under a post or answers under a question. With this in mind, we focus on the AI text detection task within this sequential text context, assuming that inputs are strictly ordered. Based on that, we develop a semantic-enhanced framework for detecting LLM-generated text (SEFD) to improve the detector's performance against paraphrasing attacks. As shown in Fig.~\ref{fig:simpleflow}, our framework comprises a retrieval pool (colored by purple) and three detection steps (colored by green). \textit{Initial detection}, which employs an existing detector to analyze the input text and generate a detection score; \textit{Semantic Similarity Computation}, a pre-trained model (e.g., BERT-based) computes semantic similarity with texts in the retrieval pool, getting a similarity score; \textit{Semantic Enhanced Detection}, a fusion function integrates the detection and similarity scores for the final score. The retrieval pool is updated using both detection and similarity scores to adapt to new LLM-generated text, enhancing paraphrase detection.
Our \textbf{contributions} can be summarized as follow:
% \begin{itemize}
%     \item We fully utilize the information about the order in which text is detected in some specific scenarios and based on this information we build a retrieval pool that is memory-efficient and can be updated in real time.
%     \item We effectively incorporate the semantic similarity retrieval technique with different principle based detector and significantly improve the detection performance under paraphrasing attack.
%     \item {\color{magenta}We provide extensive experiments across different LLM-generated texts and different detectors, demonstrating our framework's generosity and robustness against paraphrasing attacks.}
% \end{itemize}

\begin{itemize}
    \item We introduce a novel semantic-enhanced framework for detecting LLM-generated text (SEFD) that integrates retrieval-based approaches with traditional detection methods, significantly improving robustness against paraphrasing attacks.
    \item We develop an efficient and adaptive retrieval pool mechanism that balances comprehensive coverage with computational practicality, allowing real-time updates to adapt to new LLM-generated content.
    \item We demonstrate the framework's effectiveness in sequential text scenarios common in real-world applications, and provide extensive experimental validation across various LLM-generated texts and detection methods.
\end{itemize}
The rest of this paper is organized as follows: Section II reviews related work in LLM-generated text detection. Section III details our proposed SEFD framework. Section IV presents our experimental setup and results. Finally, Section V discusses the limitations of our approach, outlines future research directions, and concludes the paper.

\section{Related Work}
\noindent \textbf{Watermarking Techniques}. Current LLM-generated text detection methodologies fall into two categories: watermarking and non-watermarking. The studies \cite{zhao2023provable,zhang2023remark,he2022cater,he2022protecting,yoo2023robust,munyer2023deeptextmark,yang2023watermarking,kirchenbauer2023watermark}  leveraging watermarking approach to address the detection issues involves subtly altering the generated text in a way that is undetectable to human readers but can be identified by specialized algorithms during a post-generation analysis. Effective watermarks are meticulously designed to be resistant to removal and to exert minimal impact on the overall quality of the text output. As an example, soft watermarking developed by \cite{kirchenbauer2023watermark}, utilizes a novel token partitioning strategy, categorizing tokens into ``green'' and ``red'' lists. This partitioning aids in the creation of distinct watermark patterns, with a watermarked LLM typically selecting tokens from the ``green list'', which is determined by the preceding token, with a high degree of probability. In a recent development, Scott Aaronson~\cite{aaronsonwater} has announced his investigation into cryptographic methods of watermarking, in a collaborative effort with OpenAI. Their preliminary method is based only on biasing of the LLM output, diverging from the more deterministic approach seen in the work of \cite{fang2017generating}. While comprehensive details of this method are yet to be disclosed, preliminary information hints at the involvement of hashing techniques applied to n-gram sequences. It is important to note, however, that watermarking approaches are inherently targeted and are thus applicable only to specific LLMs. Moreover, the advancement of watermarking algorithms is somewhat constrained by the necessity for access to open-source LLMs.

\noindent \textbf{Non-watermarking Techniques}. Compared to watermarking-based methods, non-watermarking techniques offer the distinct advantage of being capable of detecting text generated by various LLMs without necessitating modifications to the generative algorithms. Early non-watermark detection strategies focused on identifying statistical anomalies in metrics such as entropy \cite{lavergne2008detecting} and perplexity \cite{beresneva2016computer}. A notable advancement in this field was the introduction of the GLTR visualizer~\cite{gehrmann2019gltr}, designed to aid human evaluators in distinguishing LLM-generated text. The emergence of ChatGPT led to the development of zero-shot detectors \cite{mitchell2023detectgpt, solaiman2019release, ippolito2019automatic, tulchinskii2024intrinsic, yang2023dna, mao2024raidar}, leveraging the statistical and topological properties of the LLM-generated text. Instead of these, classifier-based methods train supervised models to distinguish human-written text from LLM-generated text~\cite{bhattacharjee2024eagle, hovy2016enemy,zellers2019defending,bakhtin2019real,ippolito2019automatic,openaiclassifier}. 
\section{Method}
% We first formally define our detection task. For text sequence $x_1, x_2,\dots, x_N \in \Omega$, where $\Omega$ is the text space. We define $\Omega = \Omega_\textnormal{H} \cup \Omega_{f_\textnormal{LM}}$ $\Omega_\textnormal{H}$ is human-written text space and $\Omega_{f_\textnormal{LM}}$ is the text space generated by a LLM API, and $\Omega_\textnormal{H} \cap \Omega_{f_\textnormal{LM}} = \varnothing$. Our detection task is then...

We define our detection task as a binary classification problem.
Let $\texttt{X} = (x_i)_{i=1}^N$ be a text sequence, where $\texttt{X} \in \Omega$, and $\Omega$ is the text space. We define:
\begin{enumerate}
    \item $\Omega = \Omega_\textnormal{H} \cup \Omega_{f_\textnormal{LM}}$.
    \item $\Omega_\textnormal{H}$ is the human-written text space.
    \item $\Omega_{f_\textnormal{LM}}$ is the text space generated by an LLM $f_\textnormal{LM}$.
    \item $\Omega_\textnormal{H} \cap \Omega_{f_\textnormal{LM}} = \varnothing$.
\end{enumerate}
Our detection task is to define a classification function $D: \Omega \mapsto \{0, 1\}$ such that for each $x_i$:
\begin{equation}
D(x_i) = \begin{cases} 
0 & \text{if } x_i \in \Omega_\textnormal{H} \\
1 & \text{if } x_i \in \Omega_{f_\textnormal{LM}}.
\end{cases}
\end{equation}
% Where:
% \begin{itemize}
%     \item $D(x_i) = 0$ indicates that the text $x_i$ is classified as human-written
%     \item $D(x_i) = 1$ indicates that the text $x_i$ is classified as generated by the LLM $f_\textnormal{LM}$.
% \end{itemize}
% In particular, if $y_i\in\Omega_{f_{LM}}$ and $\hat{y_i}$ is $y_i$'s paraphrased version by another LLM, we still consider $\hat{y_i}\in\Omega_{f_{LM}}$. In detection phase, $\hat{y_i}$ should be classified as LLM $f_\textnormal{LM}$-generated instead of human-written.
In particular, for a text $y_i$, $\hat{y_i}$ is its paraphrased version by any LLM model. Since we only want to distinguish whether a text is human-written or not, we consider $\hat{y_i}\in\Omega_{f_{LM}}$ for simplicity, which means in the detection phase, $\hat{y_i}$ should be classified as LLM-generated instead of human-written.

%This formalization presents the detection task as a binary classification problem, where we aim to distinguish between human-written text and LLM-generated text.
Fig.~\ref{fig:flow} illustrates the structure of our framework which is a detailed version of Fig.~\ref{fig:simpleflow}. In the following sections, we detail our detection framework, starting with the three detection steps, followed by an introduction to the retrieval pool and its updating rules.
\subsection{Initial Detection} \label{initial detection}
In the initial detection phase, we employ a pre-studied detector capable of analyzing input text and generating a detection score, as presented in Step I of Alg.~\ref{alg:detection}. Mathematically, this detector can be represented as a function $f: \Omega \mapsto \mathbb{R}$, mapping from the text space to the real number. Formally, for a candidate text $x_i$, the detection score is given by:
\begin{equation}
    s_\textnormal{det}^i = f(x_i)
\end{equation}
Detectors that meet our requirements should produce distinct score distributions for different text sources, enabling differentiation between them. For instance, in soft watermarking applications, the watermarked text (LLM-generated) scores are typically significantly higher than those for human-written text.
\begin{figure}[t]
  \centering
  \includegraphics[width=8.3cm]{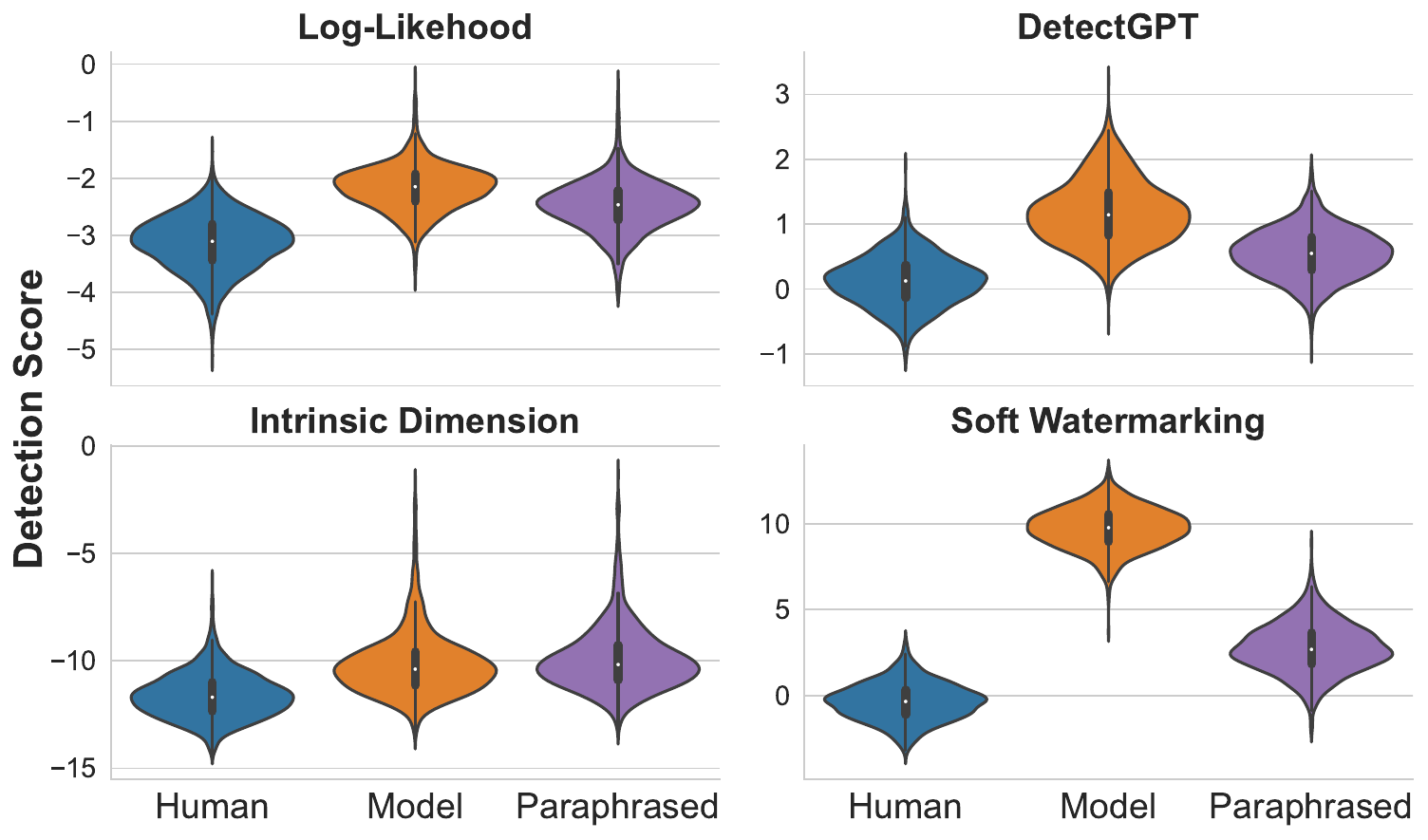}
  \caption{Detection score distributions for four different detectors: Log-Likelihood~\cite{solaiman2019release}, DetectGPT~\cite{mitchell2023detectgpt}, Intrinsic Dimension~\cite{tulchinskii2024intrinsic}, and Soft Watermarking~\cite{kirchenbauer2023watermark}. The text data consists of answers to questions from the r/explainlikeimfive subreddit. There are three groups of answers: Human for human-written answers, Model for answers generated by the GPT-2 XL model~\cite{radford2019language}, and Paraphrased for paraphrased version of GPT-2 XL generated answers using DIPPER~\cite{krishna2023paraphrasing}.}
  \label{fig:histogram}
\end{figure}

The choice of detectors can vary, with different detectors yielding unique score distributions. For real-world data, we curated 1300 questions from the r/explainlikeimfive subreddit~\footnote{https://www.reddit.com/r/explainlikeimfive/} and collect answers from three different sources: original human-written answers, LLM-generated answers with GPT-2 XL model~\cite{radford2019language}, and paraphrased versions of the LLM-generated answers using DIPPER~\cite{krishna2023paraphrasing}, a widely used paraphrasing LLM. Fig.~\ref{fig:histogram} illustrates the detection score distributions of these answers from four distinct detectors: Log-Likelihood~\cite{solaiman2019release}, DetectGPT~\cite{mitchell2023detectgpt}, Intrinsic Dimension~\cite{tulchinskii2024intrinsic}, and Soft Watermarking~\cite{kirchenbauer2023watermark}. These detectors show varying abilities to distinguish between different sources of text, as indicated by the overlapping and separation of their respective score distributions. This highlights that the effectiveness of detection can vary significantly based on the chosen detector. 
Additionally, there is a noticeable drop in the detection scores for Log-Likelihood, DetectGPT, and Soft Watermarking after paraphrasing, which makes it more challenging for these detectors to accurately classify the text. It is also important to note that the selection of detectors is not limited to these four examples.

To provide better context, we briefly introduce the four initial detectors:
\begin{itemize}
    \item \textbf{Log-Likelihood}. This method estimates the probability of a given text being generated by an LLM based on its likelihood under the model's language distribution.
    \item \textbf{DetectGPT}. This method evaluates the curvature of the model’s log-likelihood surface to identify text generated by LLMs.
    \item \textbf{Intrinsic Dimension}. This method estimates the intrinsic dimension of the text based on various techniques (e.g. maximum likelihood estimation) to differentiate LLM-generated text from human-written content.
    \item \textbf{Soft Watermarking}. This method embeds subtle watermarks in generated text to help identify LLM-produced content without noticeably altering the text's readability or structure.
\end{itemize}

\subsection{Semantic Similarity Computation}
Semantic similarity is a well-studied area \cite{reimers-2019-sentence-bert, wieting2021paraphrastic, wang2016sentence, shao2017hcti, tien2019sentence}, which is often used in pool-based retrieval method. Typically, computing the semantic similarity unfolds in a two-step process: first converting text into numerical form, which is a crucial step that lays the groundwork for understanding the underlying semantics; once texts are vectorized, the similarity is calculated using metrics such as cosine similarity, Euclidean distance, Manhattan distance, etc.

\textbf{Retrieval pool and pool size.} Given an LLM $f_\textnormal{LM}$, for any given prompt $q_k$, we use this LLM to generate a response $y_k$, represented as $y_k = f_\textnormal{LM}(q_k)$, where $y_k \in \Omega_{f_\textnormal{LM}}$. Let $f_{\textnormal{enc}}: \Omega \rightarrow \mb{R}^d$ be an encoder (e.g., sentence-transformers~\cite{reimers-2019-sentence-bert}) that embeds variable-length sequences into fixed-size vectors that encapsulate their semantic content, expressed as $\mathbf{u}_k = f_{\textnormal{enc}}(y_k)$. Let $q_1,\ldots,q_M$ be the set of prompts that have been inputted into the LLM with $y_1,\ldots,y_M$ as their output. We set the initialized pool size to be $M_0$, where $0 \leq M_0 \leq M$, and construct the retrieval pool $\mathcal{Y} = \{\mathbf{u}_1,\ldots,\mathbf{u}_{M_0}\}$ by encoding $M_0$ LLM responses with the specified encoder. Note that different LLMs may generate varying responses even with the same prompt, so it is necessary to create separate pools for each LLM.
\begin{figure}[t]
  \centering
  \includegraphics[width=8.3cm]{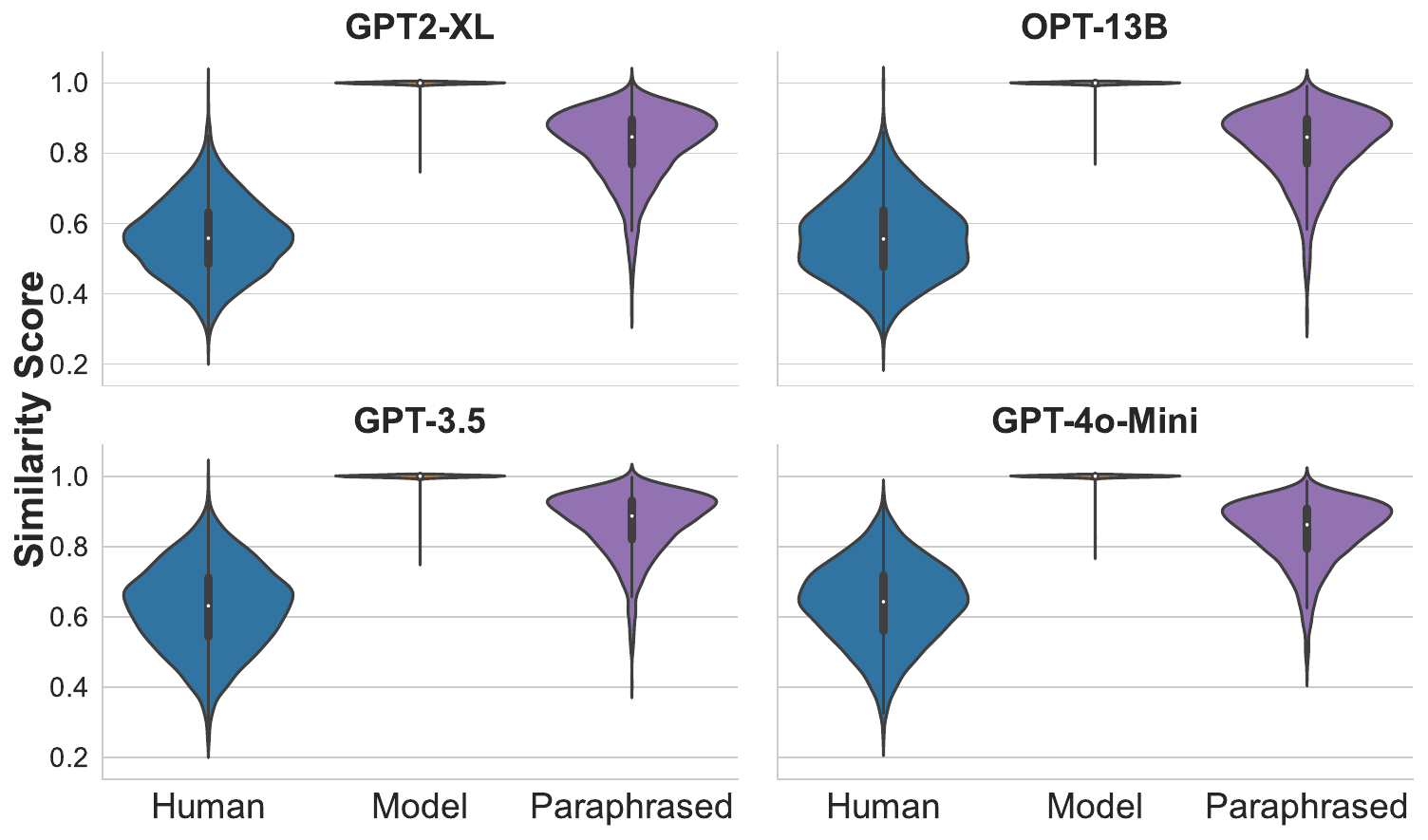}
  \caption{Similarity score distribution on four different LLM generated datasets: GPT-2 XL model, OPT-13B model, GPT-3.5 model, and GPT-4o-mini model. For each dataset, the answers come from three sources: Human for human-written answers, Model for answers generated by LLM, and Paraphrased for the paraphrased version of LLM-generated answers using DIPPER. The scores for LLM-generated answers are concentrated around 1, and the scores for paraphrased answers are obviously higher than those for human-written text.}
  \label{fig:violin}
\end{figure}

After defining the retrieval pool, for each given text, we compare its embedding with all the others in the retrieval pool to compute the semantic similarity. Specifically, let $x_i$ be a candidate text and $\mathbf{v}_i = f_{\textnormal{enc}}(x_i)$ be the vector after embedding, the maximum cosine similarity between $\mathbf{v}_i$ and $\mathbf{u}_j$ over all $\mathbf{u}_j\in \mathcal{Y}$ is:
\begin{equation}
    s_\textnormal{sim}^i = \max_{\mathbf{u}_j\in \mathcal{Y}}\frac{\langle\mathbf{v}_i,\mathbf{u}_j\rangle}{\|\mathbf{v}_i\|\|\mathbf{u}_j\|}.
\end{equation}
If $s_{\text{sim}}^i$ is close to 1, then there is a high probability that $x_i$ is generated by this LLM. Assuming the retrieval pool is large enough to cover all the distribution of the texts generated by the LLM (e.g., $M_0=M$), the similarity score of the LLM-generated text should always be 1. For better illustration, we generate answers to the 1300 questions mentioned in Section~\ref{initial detection} using four different LLMs: GPT-2 XL model~\cite{radford2019language}, OPT-13B model~\cite{zhang2022opt}, GPT-3.5 model~\cite{brown2020language}, and GPT-4o-mini~\cite{gpt40mini} model, these answers are then paraphrased using DIPPER~\cite{krishna2023paraphrasing}. For each LLM, we initialize the retrieval pool that includes all the original answers. As shown in Fig.~\ref{fig:violin}, the similarity scores of LLM-generated text are concentrated around 1. More importantly, \textit{the similarity score will still be high even if the LLM-generated text is paraphrased by another LLM, since the semantic of the text will remain essentially unchanged after paraphrasing}, whereas the similarity scores for human-written text are much lower. This distinction makes the semantic similarity retrieval technique a robust method for defending against paraphrasing. The procedure for computing semantic similarity is summarized in Step II of Alg.~\ref{alg:detection}.

As highlighted previously, achieving optimal detection results necessitates a considerably large retrieval pool, a requirement that stands as a significant limitation of this technique. This challenge is further compounded by the escalating usage of LLMs across various domains. The considerable magnitude of $M_0$ will lead to high computing requirements for executing similarity searches. While efficient methodologies such as FAISS \cite{johnson2019billion} provide some relief by utilizing adept nearest neighbor libraries, they still fall short in terms of scalability. If we use a relatively small pool to save storage and time, it may not cover all the distribution and therefore fails to provide an effective similarity score. In Section~\ref{pool rule} we delineate our pool updating rule to address this issue.

\subsection{Semantic Enhanced Detection} \label{semantic enhanced detection}
Initial detection (Step I) distinguishes itself by obviating the necessity for a training set or a similar resource pool. This characteristic is particularly beneficial in offsetting the substantial pool capacity demands of the semantic similarity computation (Step II). On the opposite, Step II demonstrates a remarkable resilience to paraphrase attacks, a feature that effectively compensates for Step I's vulnerability in similar scenarios.

\textbf{Why combine two scores?} To effectively integrate the detection score and similarity score, we observe that, as a result of the diminished capacity of the retrieval pool, a low similarity score typically indicates that the input is either human-generated or from an unincorporated LLM response. Under such conditions, the decision-making capability of Step II becomes less reliable, potentially exerting a significant influence on the judgment of initial detection. Consequently, we prioritize the insights from Step I when dealing with a lower similarity score. On the other hand, a similarity score close to 1 provides robust evidence of the input being LLM-generated, necessitating an increased dependence on Step II's assessment.

\textbf{Fusion function.} For the following discussion, we focus on detectors where human-written text has a lower detection score compared to LLM-generated text\footnote{For detectors with an inverse score distribution, one can simply use the inverse of the score.} and use min-max to normalize the detection score to $[0,1]$ before integration (in the following discussion of this part, we still use $s_\textnormal{det}$ to denote the normalized detection score). Based on the above observations and analysis, we design the following fusion function. Given parameters $\lambda_1, \lambda_2 > 0$, the \textit{fusion function} is defined as 
\begin{equation}
    f_{\textnormal{fus}}(s_\textnormal{det},s_\textnormal{sim}) = \frac{s_\textnormal{det}}{(1+10^{-\lambda_1}-s_\textnormal{sim})^{1/\lambda_2}}
\end{equation}

In the fusion function $f_{\textnormal{fus}}(s_\textnormal{det},s_\textnormal{sim})$, the coefficient $\frac{1}{(1+10^{-\lambda_1}-s_\textnormal{sim})^{1/\lambda_2}}$ serves as the weighting factor of $s_\textnormal{det}$. The parameters $\lambda_1$ and $\lambda_2$ control how strongly $s_\textnormal{sim}$ influences $s_\textnormal{det}$. In this configuration, when $s_\textnormal{sim}$ is close to 1, $s_\textnormal{det}$ will be amplified by approximately $10^{\lambda_1/\lambda_2}$ times. This allows a candidate to be classified as LLM-generated even with a small $s_\textnormal{det}$. Conversely, when $s_\textnormal{sim}$ is near 0, $s_\textnormal{det}$ is only slightly adjusted, placing greater reliance on $s_\textnormal{det}$ for the final classification outcome. However, it is important to note that, since the distribution of $s_\textnormal{det}$ is various for different detectors, the parameters $\lambda_1, \lambda_2$ need to be tuned accordingly.

With the definition of the fusion function $f_\textnormal{fus}$, we can integrate the detection score and similarity score. For a candidate text $x_i$ with its detection score $s_\textnormal{det}^i$ and similarity score $s_\textnormal{sim}^i$, the semantic enhanced detection score is:
\begin{equation}
    s_i = f_{\textnormal{fus}}(s_\textnormal{det}^i,s_\textnormal{sim}^i)
\end{equation}
With these results, we can determine if the candidate text $x_i$ is generated by the LLM $f_\textnormal{LM}$ by simply thresholding the semantic enhanced detection score $s_i$. This detection process is summarized in Step III of Alg.~\ref{alg:detection}.

\subsection{Retrieval Pool Updating Rule} \label{pool rule}
\begin{table}[t]
\centering
  \caption{The updating rule of retrieval pool.}
  \label{table: update}
  \begin{tabular}{ccccc}
    \toprule
    \textbf{Index}&$s_\textnormal{det}$&$s_\textnormal{sim}$& \textbf{Add}& \textbf{Replace}\\
    \midrule
    Situation 1 & $\geq\epsilon_\textnormal{det}$ & $\geq\epsilon_\textnormal{sim}$ & \XSolidBrush & --\\
    \midrule
    Situation 2 & $\geq\epsilon_\textnormal{det}$ & $< \epsilon_\textnormal{sim}$& \Checkmark & \XSolidBrush\\
    \midrule
    Situation 3 & $< \epsilon_\textnormal{det}$ & $\geq\epsilon_\textnormal{sim}$ & \Checkmark & \Checkmark\\
    \midrule
    Situation 4 & $< \epsilon_\textnormal{det}$ & $< \epsilon_\textnormal{sim}$ & \XSolidBrush & --\\
  \bottomrule
\end{tabular}
\end{table}

\textbf{Input text assumptions.} 
%\hbj{We need to first emphasize this assumption is very common in real-world applications, for example: According to the key observation on the sequential text in online forums, social media platforms, question-answer websites etc. where the responses usually appear in sequential form and the paraphrased texts usually appear after the original texts, we summarize the following input text assumption:} 
Many real-world applications involve sequential text, where accurate detection is important. For example, in online forums or social media platforms like X and Instagram, comments are chronologically arranged under each post and determining whether a comment is generated by a user or an automated system is key to effective content moderation. Also, on question-answer websites like Stack Overflow and Quora, where users submit questions and receive answers in a sequential format, it is crucial to determine whether responses are from LLM or are original human-generated content. This distinction helps maintain the integrity of the information and prevents the spread of misleading content. Educational platforms and customer support systems, which also utilize sequential text, similarly benefit from robust detection mechanisms to uphold content accuracy and relevance. In this paper, we mainly focus on the LLM-generated text detection task in this sequential text scenario. Based on the observation, we first define \textit{input text assumptions} for the input text sequence $\texttt{X} = (x_i)_{i=1}^N$:
\begin{quote}
    \textbf{A. 1}: The inputs are in a strictly sequential order, which means two or more texts cannot be input at the same time. \\
    \textbf{A. 2}: For an LLM-generated text, the paraphrased version always comes after the original text.\\
\end{quote}
For example, let “$a$,” “$b$,” and “$c$” represent LLM-generated texts, and “$\hat{a}$,” “$\hat{b}$,” and “$\hat{c}$” represent their paraphrased versions. A valid input sequence would be: “$a;b;\hat{a};\hat{b};c;\hat{c}$.”

These two assumptions are important for updating the retrieval pool effectively. A. 1 ensures that we can examine each text sequentially, one at a time, maintaining the order of input. A. 2 is fundamental to our approach, as it stipulates that any paraphrased LLM-generated text will only appear after its original LLM-generated text. This sequential ordering is essential for defending against paraphrasing attacks. The logic behind this assumption is intuitive: to paraphrase an LLM-generated text using another LLM, the original text must already exist. Next, we introduce the updating rule.

\textbf{Updating rule}. For each input candidate text $x_i \in \texttt{X}$, we decide whether to add it into the retrieval pool or not based on the updating rule. We first define two thresholds $\epsilon_\textnormal{det}$ and $\epsilon_\textnormal{sim}$. Given the detection score $s_\textnormal{det}^i$ and similarity score $s_\textnormal{sim}^i$ of the text $x_i$, the updating rule is presented in Table~\ref{table: update} and the process is summarized in Alg.~\ref{alg:detection}. There are four situations presented in Table~\ref{table: update}, for the sake of narrative we define the text in the retrieval pool\footnote{Our retrieval pool stores the embedding of text, not the text itself.} that has the highest similarity score with $x_i$ as $y_{\tau}$, then we explain the details as follow:
\begin{figure}[t]
  \centering
  \includegraphics[width=7cm]{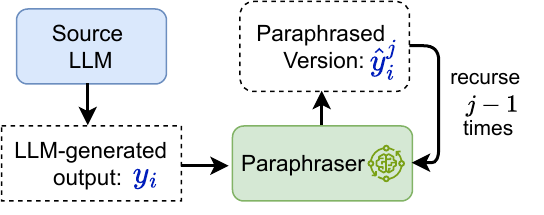}
  \vspace{-0.3cm}
  \caption{Recursive paraphrasing}
  \label{fig:recursive}
\end{figure}
\begin{itemize}
    \item \textit{For Situation 1}, the input $x_i$ has a high value in both detection score and similarity score, which shows strong evidence that $x_i$ is from LLM. According to A. 2, $x_i$ is either already in the pool  ($x_i = y_{\tau}$) or is a minor paraphrase of $y_{\tau}$ (as a minor paraphrase can still result in a high detection score, even though paraphrasing generally lowers it, as shown in Fig.~\ref{fig:histogram}). Furthermore, since the initial detection successfully detects $x_i$, we opt not to add $x_i$ into the pool in order to save space.
    \item \textit{For Situation 2}, the scores of $x_i$ show that $x_i$ is generated by LLM but it is not semantically similar to any text in the pool, which means $x_i$ is from LLM and it is not a paraphrased version. We add it into the pool prepared for the future detection of its paraphrased version.
    \item \textit{For Situation 3}, high similarity score shows strong evidence that $x_i$ is either already in the pool ($x_i = y_{\tau}$) or a paraphrase of $y_{\tau}$, but the low detection score indicates it evades the initial detection. In both cases, we opt to replace $y_{\tau}$ with $x_i$ to update the pool while controlling the number of texts in the pool. If $x_i$ is already in the pool, $x_i$ is identical to $y_{\tau}$ and this update will not change the pool. If $x_i$ is a paraphrased version of $y_{\tau}$, this replacement helps defend against \textit{recursive paraphrasing attacks} (paraphrasing a text more than once). More specifically, while semantic similarity remains high after a single paraphrase, it tends to decrease when the paraphrase performs in recursion~\cite{sadasivan2023can} as shown in Fig.~\ref{fig:recursive}. With $x_i$ added into the pool, when detecting its paraphrased version, denoted as $\hat{x}_i^1$, although the semantic similarity between $\hat{x}_i^1$ and $y_{\tau}$ may significantly decrease, $\hat{x}_i^1$ maintains high similarity with $x_i$, so the similarity score remains high and $\hat{x}_i^1$ will be added into the pool to replace $x_i$. This process continues for detecting $\hat{x}_i^2, \hat{x}_i^3,\cdots$.
    \item \textit{For Situation 4}, both detection score and similarity score don't show evidence that $x_i$ is from LLM, so we choose not to add it to the pool.
\end{itemize}
Two key factors can influence the effectiveness of the update process. The first is the choice of threshold $\epsilon_\textnormal{det}$. Since different detectors produce varying detection score distributions, $\epsilon_\textnormal{det}$ must be adjusted to suit each detector accordingly. The second factor is the performance of the initial detection. Ideally, we aim for a high detection score for LLM-generated text and a low detection score for human-written text. However, this ideal scenario is not always achieved with the initial detectors, which may result in some LLM-generated texts being missed and some human-written texts being mistakenly added to the pool during the updating process.
\\

In summary, our detection framework operates as follows: For a candidate text $x_i\in\texttt{X}$, we conduct initial detection and semantic similarity computation simultaneously to get detection score and similarity score. Then we apply the fusion function to integrate these two scores and produce the semantic enhanced detection score to classify $x_i$. Finally, based on the updating rule, we determine whether to incorporate $x_i$ into the pool and continue to detect the next candidate text $x_{i+1}$. The complete workflow of our detection framework is presented in Alg.~\ref{alg:detection}.

\begin{algorithm}[t]
\small
\caption{Semantic-Enhanced Framework for Detecting LLM-Generated Text (SEFD)}\label{alg:detection}
\begin{algorithmic}[1]
\Require 
input text sequence $\texttt{x} = (x_i)_{i=1}^N \in \Omega $ (where $\Omega$ is the text space), detector $f: \Omega\mapsto\mb{R}$, encoder function $f_{\textnormal{enc}}:\Omega\mapsto\mb{R}^d$, the size of retrieval pool $M_0$, retrieval pool $\mathcal{Y} = \{\mathbf{u}_1,\ldots,\mathbf{u}_{M_0}\}$, fusion function $f_{\textnormal{fus}}: \mb{R}\times\mb{R}\mapsto\mb{R}$, initial detection threshold $\epsilon_{\text{det}}$, semantic similarity threshold $\epsilon_{\text{sim}}$, decision threshold $\epsilon$.

\Statex \algorithmicline
\vspace{0.05cm}
\noindent \centerline{\textbf{Step~I: Initial Detection}}
\State $s_{\textnormal{det}}^i \gets f(x_i)$, $s_{\textnormal{det}}^i \in \mb{R}$

\Statex \algorithmicline
\vspace{0.05cm}
\noindent \centerline{\textbf{Step~II: Semantic Similarity Computation}}
\centerline{(conduct simultaneously with Step I)}
\State $\mathbf{v}_i \gets f_{\textnormal{enc}}(x_i)$, $\mathbf{v}_i\in \mb{R}^d$
\For {$j = 1, \ldots, M_0$}
    \State $\eta_{i}^j = \frac{\langle\mathbf{v}_i,\mathbf{u}_j\rangle}{\|\mathbf{v}_i\|\|\mathbf{u}_j\|}$
\EndFor
\State $s_{\textnormal{sim}}^i \gets \eta_i^{\tau} = \max \left\{\eta_i^1, \eta_i^2, \ldots, \eta_i^{M_0} \right\}$, $s_{\textnormal{sim}}^i \in \mb{R}$

\Statex \algorithmicline
\vspace{0.05cm}
\noindent\centerline{\textbf{Step~III: Semantic Enhanced Detection}}
\State $s_i\gets f_{\textnormal{fus}}(s_{\textnormal{det}}^i,s_{\textnormal{sim}}^i)$
\If{$s_i > \epsilon$}
    \State \Return 1 (from LLM)
\Else
    \State \Return 0 (from human)
\EndIf

\Statex \algorithmicline
\vspace{0.05cm}
\noindent \centerline{\textbf{{Pool Update}}}
\If{$s_{\text{det}}^i \geq \epsilon_{\text{det}}$ \textbf{and} $s_{\text{det}}^i \geq \epsilon_{\text{sim}}$} $\mathcal{Y} \gets \mathcal{Y}$
\ElsIf{$s_{\text{det}}^i < \epsilon_{\text{det}}$ \textbf{and} $s_{\text{det}}^i < \epsilon_{\text{sim}}$} $\mathcal{Y} \gets \mathcal{Y}$
\ElsIf{$s_{\text{det}}^i \geq \epsilon_{\text{det}}$ \textbf{and} $s_{\text{det}}^i < \epsilon_{\text{sim}}$} $\mathcal{Y} \gets \mathcal{Y} \cup \{\mathbf{v}\}$
\ElsIf{$s_{\text{det}}^i < \epsilon_{\text{det}}$ \textbf{and} $s_{\text{det}}^i \geq \epsilon_{\text{sim}}$}
 $\mathcal{Y} \gets \mathcal{Y}\setminus  \{\mathbf{u}_{\tau}\} \cup \{\mathbf{v}\}$
\EndIf

\end{algorithmic}
\end{algorithm}

\section{Experiments}
We conduct experiments to gain deeper insights into various aspects of detecting LLM-generated text. We study the effectiveness of SEFD for detecting LLM-generated text and defending against paraphrasing attacks across various datasets. To further understand the impact of the retrieval pool, we also examine the detection accuracy under different initial pool sizes. Finally, as an additional demonstration of our method's robustness, we evaluate SEFD's performance against recursive paraphrasing attacks, which can severely degrade the effectiveness of traditional detectors.
\subsection{Evaluation Metrics}
In our experiment, we use two metrics to evaluate the performance of detection.

\textbf{AUROC} Since detection is fundamentally a binary classification task, the results are largely dependent on the chosen decision threshold ($\epsilon$ in Alg.~\ref{alg:detection}). The first metric is the AUROC (Area Under the Receiver Operating Characteristic Curve) which is commonly used to measure detection performance \cite{mitchell2023detectgpt,krishna2023paraphrasing}, assessing detection performance across the spectrum of potential thresholds. 

\textbf{Detection Accuracy} Typically, we anticipate the detection process to exhibit a high true positive rate (TPR) performance. However, in the detection of LLM-generated text, maintaining a low false positive rate (FPR) is also significant; in other words, human-written text must be rarely misclassified as LLM-generated \cite{kirchenbauer2023watermark,sadasivan2023can}. This requirement is straightforward and intuitive since an AI language detector without a low FPR can cause harm as it might wrongly accuse a human of plagiarizing using an LLM. Therefore, by adjusting the detection threshold, we compute the true positive rate (TPR) at 1\% FPR, denoted as \textit{detection accuracy}, which serves as our second metric. Compared with AUROC, detection accuracy is more important in real scenario detection, so in our experiment, we will focus more on detection accuracy.

\subsection{Baselines, Data, and Settings}
\textbf{Base language models} In our experiment, we want to assess and compare the performance of detectors in identifying text produced by different LLMs. We focus on four base language models: (1) GPT-2 XL model~\cite{radford2019language}, which possesses 1.5B parameters; (2) OPT-13B model~\cite{zhang2022opt}, renowned for its unique architecture; (3) {\fontfamily{qcr}\selectfont text-davinci-003} variant from GPT-3.5 family~\cite{brown2020language}, which has 175B parameters and has additionally been instruction tuned using reinforcement learning from human feedback (RLHF)~\cite{ouyang2022training}; (4) GPT-4o mini~\cite{gpt40mini}, a latest cost-efficient version of GPT-4 with only 8B parameters.

\textbf{LLM-generated text data} 
Our study explores long-form text generation tasks, primarily due to their association with potentially harmful uses, such as the fabrication of false articles. We specifically focus on long-form question answering. This involves a language model providing detailed responses of 250-300 words to complex how/why queries, like ``Why did it take so long for sunglasses to get widespread?''. To create a relevant dataset for this task, we extract questions from Reddit, targeting six prominent domains: biology, physics, chemistry, economics, law, and technology. From each domain, 500 questions are randomly selected. Each question is paired with its most comprehensive human-written response found on the subreddit, resulting in 3,000 long-form question-answer pairs. These questions are then used as prompts for the four aforementioned base language models—GPT-2 XL, OPT-13B, GPT-3.5, and GPT4o-mini—to respectively generate 3,000 responses labeled as LLM-generated text.

\begin{table}[t]
\centering
  \caption{The parameters settings under each initial detector}
  \label{table: para}
  \begin{tabular}{ccccc}
    \toprule
    \textbf{Initial detector}&$\epsilon_{\text{det}}$ & $\epsilon_{\text{sim}}$ & $\lambda_1$ & $\lambda_2$\\
    \midrule
    Log-likelihood & -2.5 & 0.85 & 1 & 6\\
    \midrule
    DetectGPT & 0.5 & 0.85 & 1 & 6\\
    \midrule
    Intrinsic Dimension & -11 & 0.85 & 1 & 6\\
    \midrule
    Soft Watermarking & 4 & 0.85 & 1 & 6\\
  \bottomrule
\end{tabular}
\end{table}

\textbf{Paraphrasing LLM-generated text data} Paraphrasing presents a significant challenge to detectors, often diminishing their effectiveness, as evidenced by reduced AUROC scores and detection accuracy~\cite{krishna2023paraphrasing}. In our experiment, we assess the resilience of models when faced with paraphrasing attacks. To this end, we utilize DIPPER~\cite{krishna2023paraphrasing}, a paraphrasing model with 11 billion parameters. DIPPER is able to handle context in the form of prompts or multi-sentence inputs and keep the input semantics; in other words, given a prompt (such as a question) and an input text (the answer), DIPPER is capable of producing text that conveys a similar meaning to the original input text (the answer).
We input the questions as prompts along with the LLM-generated responses from four base language models into DIPPER for paraphrasing. After that, for each question under every base language model, we obtain three versions of answers: one from a human, one directly from the LLM, and one paraphrased by DIPPER.

\textbf{Input data summary} We now have four datasets, each containing three versions of answers to every question. To ensure the input text sequence aligns with our input text assumptions during detection, we structure the input text sequence for each dataset as follows: first, we input all the LLM-generated texts, followed by the human-written answers, and finally, the paraphrased versions.

\begin{table*}[t]
\centering
\setlength{\tabcolsep}{3pt}
\caption{Detection Performance of different methods on various datasets, measured by AUROC and detection accuracy (DA). Arrows (↑) indicate improved performance. *GPT-4o ID-MLE uses TPR at 10\% FPR, as it scores 0\% at 1\% FPR.}
\label{table: results}
\begin{tabular}{lcccccccc}
\toprule
\textbf{Dataset} & \multicolumn{2}{c}{\textbf{GPT-2 XL}} & \multicolumn{2}{c}{\textbf{OPT-13B}} & \multicolumn{2}{c}{\textbf{GPT-3.5}} & \multicolumn{2}{c}{\textbf{GPT-4o mini}} \\
\cmidrule(lr){2-3} \cmidrule(lr){4-5} \cmidrule(lr){6-7} \cmidrule(lr){8-9}
\textbf{Detect Method} & \textbf{original} & \textbf{paraphrased} & \textbf{original} & \textbf{paraphrased} & \textbf{original} & \textbf{paraphrased} & \textbf{original} & \textbf{paraphrased} \\
\textbf{Metrics} & AUROC / DA & AUROC / DA & AUROC / DA & AUROC / DA & AUROC / DA & AUROC / DA & AUROC / DA & AUROC / DA \\
\midrule
Likelihood & 0.95 / 49.4 & 0.85 / 19.4 & 0.89 / 29.2 & 0.83 / 17.8 & 0.97 / \textbf{71.2} & 0.93 / 40.0 & 0.86 / 18.4 & 0.82 / 10.8 \\
+ pool size -0 & 0.93/ 38.0 & 0.88 / 40.7 ↑ & 0.87 / 23.5 & 0.85 / \textbf{34.8} ↑ & 0.96 / 56.0 & 0.95 / 59.9 ↑ & 0.84 / 11.0 & 0.83 / 22.1 ↑ \\
+ pool size -1/5 & 0.94 / \textbf{50.5} ↑ & 0.89 / \textbf{40.9} ↑ & 0.89 / \textbf{32.7} ↑ & 0.85 / 33.9 ↑ & 0.96 / 59.7 & 0.95 / \textbf{63.0} ↑ & 0.85 / \textbf{24.4} ↑ & 0.84/ \textbf{30.0} ↑ \\
\midrule
DetectGPT & 0.96 / 58.54 & 0.79 / 8.72 & 0.86 / 13.96 & 0.76 / 4.46 & 0.91 / 26.99 & 0.82 / 7.41 & 0.82 / 9.69 & 0.75 / 4.46 \\
+ pool size -0 & 0.96 / 54.56 & 0.83 / 25.85 ↑ & 0.85 / 11.68 & 0.80 / 15.45 ↑ & 0.90 / 20.95 & 0.86 / 24.89 ↑ & 0.81 / 8.53 & 0.78 / 13.29 ↑ \\
+ pool size -1/5 & 0.97 / \textbf{64.31} ↑ & 0.83 / \textbf{26.06} ↑ & 0.87 / \textbf{28.77} ↑ & 0.80 / \textbf{15.66} ↑ & 0.92 / \textbf{36.46} ↑ & 0.86 / \textbf{25.15} ↑ & 0.84 / \textbf{25.72} ↑ & 0.78 / \textbf{13.80} ↑ \\
\midrule
ID-MLE & 0.81 / 11.0 & 0.85 / 13.8 & 0.80 / 12.0 & 0.84 / 14.7 & 0.83 / 4.1 & 0.86 / 7.5 & * 0.50 / 4.4 & * 0.69 / 21.8 \\
+ pool size -0 & 0.80 / 10.4 & 0.87 / \textbf{24.5} ↑ & 0.79 / 11.2 & 0.86 / 24.7 ↑ & 0.82 / 3.0 & 0.89 / 16.0 ↑ & * 0.51 / 4.3 & * 0.71 / 26.7 ↑ \\
+ pool size -1/5 & 0.82 / \textbf{18.0} ↑ & 0.87 / 24.0 ↑ & 0.81 / \textbf{15.1} ↑ & 0.86 / \textbf{24.8} ↑ & 0.84 / \textbf{6.7} ↑ & 0.89 / \textbf{16.2} ↑ & * 0.60 / \textbf{26.0} ↑ & * 0.75 / \textbf{43.4} ↑ \\
\midrule
Watermarking & 1.0 / 100 & 0.96 / 61.7 & 1.0 / 100 & 0.95 / 57.9 & - & - & - & - \\
+ pool size -0 & 1.0 / 100 & 0.99 / \textbf{90.7} ↑ & 1.0 / 100 & 0.98 / \textbf{75.2} ↑ & - & - & - & - \\
+ pool size -1/5 & 1.0 / 100 & 0.99 / 85.0 ↑ & 1.0 / 100 & 0.97 / 71.4 ↑ & - & - & - & - \\
\bottomrule
\end{tabular}
\end{table*}

\textbf{Initial detectors and other settings.} We experiment with four different initial detectors: Log-Likelihood~\cite{solaiman2019release}, DetectGPT~\cite{mitchell2023detectgpt}, Intrinsic Dimension~\cite{tulchinskii2024intrinsic}, and Soft Watermarking~\cite{kirchenbauer2023watermark}. For Intrinsic Dimension, we adopt the maximum likelihood estimation (MLE)~\cite{levina2004maximum} to approximate the intrinsic dimension instead of the persistent homology dimension (PHD) estimator proposed in~\cite{tulchinskii2024intrinsic}, as PHD includes random sampling, thus causing diverse unsatisfactory results and the result of MLE is more deterministic and robust. We use the default settings for other detectors. In the semantic similarity computation phase, we employ sentence-transformers~\cite{reimers-2019-sentence-bert} as our encoder, which is  a modification of the pre-trained BERT~\cite{devlin2018bert} network optimized for generating semantically meaningful sentence embeddings. As mentioned before, we have four tuning parameters: $\epsilon_{\text{det}}, \epsilon_{\text{sim}}, \lambda_1, \lambda_2$, and Table~\ref{table: para} presents the corresponding parameters settings. It is important to note that the chosen parameters may not yield the optimal results; they are set to reasonable values to present the results.

% \textbf{Metrics} & \textbf{original} & \textbf{paraphrased} & \textbf{original} & \textbf{paraphrased} & \textbf{original} & \textbf{paraphrased} & \textbf{original} & \textbf{paraphrased} \\

\subsection{Results}
\subsubsection{Main Results}
We first present how SEFD can improve the detection performance across four initial detectors: Log-Likelihood (likelihood), DetectGPT, Intrinsic Dimension with MLE (ID-MLE), and Soft Watermarking (Watermarking). The results are shown in Table~\ref{table: results}. In this table, “+ pool size -0” indicates SEFD with an empty initial pool, while “+ pool size -1/5” indicates SEFD with an initial pool containing 1/5 of the LLM-generated text.

\textbf{Effective defense against paraphrase.} From the results in Table~\ref{table: results}, we observe that SEFD significantly improves detection performance in both AUROC and detection accuracy under paraphrase attacks across all datasets and initial detectors. Intuitively, a larger pool size generally leads to better performance since when detecting text already present in the pool, the similarity score approaches~1, allowing for accurate classification. However, there are cases that SEFD performs better with an empty initial pool. This is primarily due to the lack of robustness in the semantic encoder we applied\footnote{The encoder used is the \texttt{all-MiniLM-L6-v2} from sentence-transformers~\cite{reimers-2019-sentence-bert}, which is efficient but slightly sacrifices performance.}. Given that both human-written and LLM-generated texts answer the same question, they often share common terminology related to the query. A less robust encoder may cause the semantic similarity based technique to incorrectly identify high semantic similarity between a human-written answer and an LLM-generated text stored in the initial pool, potentially leading to misclassification. In this rare situation, an empty pool can avoid such mistake and thus lead to a better performance than a non-empty pool. The influence of the initial pool size will be further discussed in Section~\ref{influence of pool size}.

% Two factors can contribute to this. First, the initial detector already performs well in detecting LLM-generated text (e.g. soft watermarking), so with our updating rule, most of LLM-generated text can be detected and added into the pool, enhancing paraphrase detection even with an empty initial pool \hbj{how does this indicate an empty pool is better?}; Second, while the initial pool helps defend against paraphrasing, it can also negatively impact the recognition of human-written texts, as they might share semantic similarities with the LLM-generated texts stored in the pool\hbj{need to be careful about this point since it may not be consistent with the previous statement: semantically similar with the AI text in the pool indicates the target text is an AI text}.
\textbf{Performance on detecting LLM-generated text.} Besides the improvement on defending paraphrase, SEFD also enhances detection of original LLM-generated text. With the initial pool size 1/5, detection accuracy shows significant improvement. However, with an empty initial pool, the detection accuracy decreases for Log-likelihood, DetectGPT, and Intrinsic Dimension. This is because, with an empty pool, the semantic similarity retrieval technique cannot assist in detecting LLM-generated text. Additionally, as the detection process continues and texts are added to the pool, the similarity score becomes helpful for detecting paraphrases but may also negatively influence the initial detection decision, as discussed in Section~\ref{semantic enhanced detection}. Notably, if the initial detector is robust enough in classifying LLM-generated text, such as watermarking, this decrease will not occur, as shown in the results of Watermarking in Table~\ref{table: results}.

\subsubsection{Influence of Initial Pool Size}  \label{influence of pool size}
\begin{figure}[t]
  \centering
  \includegraphics[width=8.3cm]{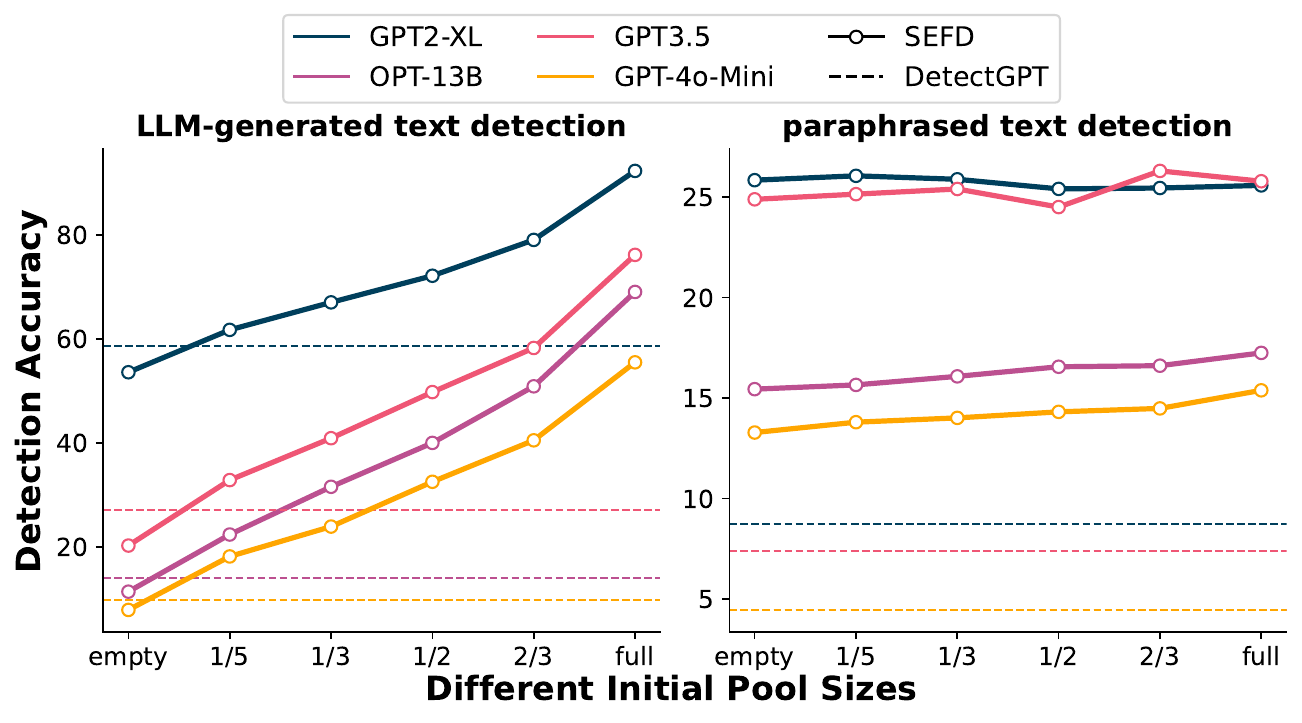}
  \vspace{-0.3cm}
  \caption{Detection accuracy under different initial pool sizes with DetectGPT~\cite{mitchell2023detectgpt} as the initial detector. The $x$-axis represents the proportion of LLM-generated text included in the initial pool. Different colors represent datasets from different LLMs. Solid lines with circle markers indicate detection performance of SEFD, while dotted lines show the results of DetectGPT.}
  \label{fig:pool size}
\end{figure}

In this section, we investigate the influence of initial pool sizes on detection performance. Fig.~\ref{fig:pool size} illustrates our findings, with DetectGPT serving as the initial detector. The $x$-axis represents the proportion of LLM-generated text included in the initial pool. A non-empty initial retrieval pool is expected to enhance both original LLM-generated text detection and paraphrase detection: For LLM-generated text detection, the semantic similarity retrieval technique can precisely identify text that are already stored in the pool, outputting a similarity score of 1 to classify the text as LLM-generated. For paraphrase detection, when detecting a paraphrased version of text stored in the pool, the similarity score, while not equal to 1, will be close to it. This score is still larger than that of human-written text, thus aiding detection.

As evident in Fig.~\ref{fig:pool size}, detection accuracy improves as the pool size increases for both detection tasks. Notabaly, the degree of improvement for paraphrased detection is less pronounced compared to that of original LLM-generated text detection. Thanks to our updating rule, even with an empty initial pool SEFD already has a significant improvement when facing paraphrasing attacks. The updating rule allows the retrieval pool to incorporate new LLM-generated text, which aids in paraphrase detection. Consequently, paraphrase detection does not solely rely on the initial pool, explaining the more modest improvement compared to original text detection.
\subsubsection{Recursive Paraphrase} \label{recursive paraphrase}
In this section, we evaluate SEFD's detection performance against recursive paraphrasing attacks. Previous work~\cite{sadasivan2023can} has demonstrated that such attacks can circumvent many detection methods, including watermarking, while only slightly degrading text quality. As discussed in Section~\ref{pool rule}, one of our updating rules is specifically designed to counter recursive paraphrasing by incorporating paraphrases into the retrieval pool.

In this experiment, we apply DIPPER~\cite{krishna2023paraphrasing} in recursion to paraphrase LLM-generated answers three times. For each question, we now have five different versions of answers: one human-authored, one LLM-generated, and three recursive paraphrases. We use DetectGPT~\cite{mitchell2023detectgpt} as the initial detector and set the initial retrieval pool size to encompass 1/5 of the LLM-generated text. The result is presented in Fig.~\ref{fig:recursive_results}, where “\text{pp}$i$” on the x-axis denotes the $i$th paraphrased version.

The results reveal a significant drop in both AUROC and detection accuracy across different datasets for DetectGPT. Notably, after three rounds of paraphrasing, DetectGPT's detection accuracy approaches zero. In contrast, while SEFD's detection performance does decrease, the decline is considerably more moderate. SEFD significantly improves detection performance across all scenarios compared to DetectGPT.
\begin{figure}[t]
  \centering
  \includegraphics[width=8.3cm]{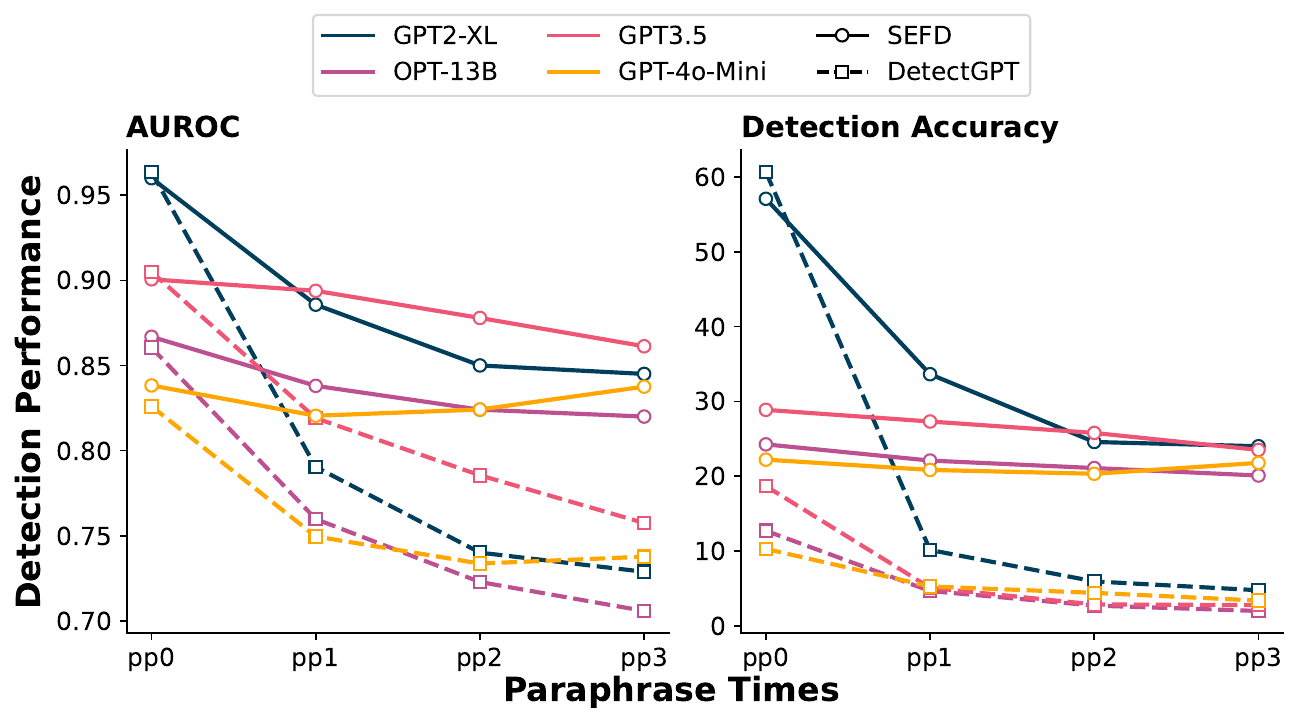}
  \vspace{-0.3cm}
  \caption{Detection performance of SEFD under recursive paraphrasing attacks. The initial detector is DetectGPT~\cite{mitchell2023detectgpt}. ``pp$i$'' on the x-axis denotes the i-th paraphrased version. Different colors represent different datasets from different LLMs. The circle markers indicate detection results from our SEFD, while the square markers represent detection results from DetectGPT.}
  \label{fig:recursive results}
\end{figure}
\section{Discussion and Conclusion} \label{discussion and conclusion}
\subsection{Limitations and Future Works}
\textbf{Limitations.} One limitation of our work is the lack of globally optimal parameters. Our semantic-enhanced detection framework employs four tuning parameters: $\epsilon_{\text{det}}, \epsilon_{\text{sim}}, \lambda_1,$ and $\lambda_2$. In our experiments, we selected workable values to demonstrate our method's efficacy. However, these parameters warrant further investigation to potentially improve detection performance. It is important to note that there is no universally optimal choice for these parameters, as their ideal values largely depend on the initial detector used. Furthermore, we discuss only one type of fusion function in this study. While effective, this represents just one possible approach. The optimal choice of fusion function may vary depending on the characteristics of different initial detectors.

\textbf{Future Work.} While our proposed framework primarily focuses on improving detection performance, future work may address equitable detection across diverse groups of human-written text. Studies have shown that GPT detectors can be biased against non-native English writers~\cite{liang2023gpt}. This fairness concern may be widespread in LLM-generated text detection and has implications for human rights. Besides, the relationship between prompting strategies and detection efficacy remains unexplored. Future work should examine whether clever prompts, such as instructing LLMs to generate answers in a more human-like tone, can successfully evade existing detection methods. Finally, the boundary between humans and AI needs to be further studied. When a human-written text is paraphrased by an LLM, classifying the paraphrase as either purely human-written or LLM-generated is not persuasive. Future research may develop more nuanced classification systems that go beyond the simple binary of human-written or LLM-generated.
\subsection{Conclusion}
% This study presents SEFD, a novel semantic-enhanced framework for detecting LLM-generated text, designed to address the growing challenge of identifying AI-generated content, particularly in the face of paraphrasing and recursive paraphrasing attacks. Our approach innovatively combines retrieval-based mechanisms with traditional detectors, leveraging text semantics to significantly improve detection performance and robustness. Through extensive experiments across various LLM-generated texts and detection methods, we have demonstrated SEFD's superior performance in both standard and paraphrasing scenarios. Importantly, the framework is highly adaptable and can be applied to different detectors. We hope this work can inspire future research to develop effective, general-purpose methods to mitigate the potential risks of AI misuse.

This study introduces SEFD, a semantic-enhanced framework for detecting LLM-generated text, addressing challenges like paraphrasing and recursive paraphrasing attacks. By integrating retrieval-based mechanisms with traditional detectors, SEFD leverages text semantics to enhance detection performance and robustness. Extensive experiments demonstrate its superior performance across various LLM-generated texts and detection methods, in both standard and paraphrasing scenarios. Notably, SEFD is highly adaptable and compatible with different detectors. We this work can inspire future research toward developing effective, general-purpose solutions to mitigate AI misuse.

\section*{Acknowledgment}

This work is supported in part by NIH Grant P30 AG073105.

%\section*{References}

\bibliographystyle{ieeetr}
\bibliography{bibliography}
\end{document}